\documentclass[10pt, a4paper]{article}
\usepackage{lrec}
\usepackage{multibib}
\newcites{languageresource}{Language Resources}
\usepackage{graphicx}
\usepackage{tabularx}
\usepackage{soul}
\usepackage{epstopdf}
\usepackage[latin1]{inputenc}
\usepackage{hyperref}
\usepackage{xstring}
\usepackage{times}
\usepackage{latexsym}
\usepackage{colortbl}
\usepackage{url}
\usepackage{graphicx}
\usepackage{tabularx}
\usepackage{soul}
\usepackage{booktabs} 
\usepackage{nicefrac}
\usepackage{latexsym}
\usepackage{multirow}
\usepackage{url}
\usepackage{array}
\usepackage{ctable}
\usepackage{epstopdf}
\usepackage[latin1]{inputenc}
\usepackage{hyperref}
\usepackage{xstring}
\usepackage{color}
\newcolumntype{P}[1]{>{\centeringarraybackslash}p{#1}}

\newcommand{\secref}[1]{\StrSubstitute{\getrefnumber{#1}}{.}{ }}

\title{Using Distributional Thesaurus Embedding for Co-hyponymy Detection}
\name{Abhik Jana, Nikhil Reddy Varimalla and Pawan Goyal}
\address{Universit{\"a}t Hamburg, Indian Institute of Technology Kharagpur, Indian Institute of Technology Kharagpur \\
        jana@informatik.uni-hamburg.de,  nikhil.varimala@gmail.com, pawang@cse.iitkgp.ac.in\\}

\abstract{
Discriminating lexical relations among distributionally similar words has always been a challenge for natural language processing (NLP) community. 
In this paper, we investigate whether the network embedding of distributional thesaurus can be effectively utilized to detect co-hyponymy relations. By extensive experiments over three benchmark datasets, we show that the vector representation obtained by applying node2vec on distributional thesaurus outperforms the state-of-the-art models for binary classification of co-hyponymy vs. hypernymy, as well as co-hyponymy vs. meronymy, by huge margins. 
\\ \newline  }

\begin{document}

\maketitleabstract

\section{Introduction}

Distributional semantic models are used in a wide variety of tasks like sentiment analysis, word sense disambiguation, predicting semantic compositionality, etc. Automatic detection of lexical relations is one such fundamental task which can be leveraged in applications like paraphrasing, ontology building, metaphor detection etc. Both supervised and unsupervised methods have been proposed by the researchers to identify lexical relations like hypernymy, co-hyponymy, meronymy etc. over the years. Recent attempts to solve this task deal with proposing similarity measures based on distributional semantic models~\cite{roller2014inclusive,weeds2014learning,SANTUS16.455,shwartz-santus-schlechtweg:2017:EACLlong,roller-erk:2016:EMNLP2016}. For hypernymy detection, several works use distributional inclusion hypothesis~\cite{geffet2005distributional}, entropy-based distributional measure~\cite{santus2014chasing} as well as several embedding schemes~\cite{fu2014learning,yu2015learning,nguyen-EtAl:2017:EMNLP2017}. Image generality for lexical entailment detection~\cite{kiela2015exploiting} has also been tried out for the same purpose. 
As far as meronymy detection is concerned, most of the attempts are pattern based~\cite{berland1999finding,girju2006automatic,pantel2006espresso} along with some recent works exploring the possibility of using distributional semantic models~\cite{morlane2015can}. 

Similarly, for co-hyponymy detection, researchers have investigated the usefulness of several distributional semantic models. One such attempt is made by~\newcite{weeds2014learning}, where they proposed a supervised framework and used several vector operations as features for the classification of hypernymy and co-hyponymy.~\newcite{SANTUS16.455} proposed a supervised method based on a Random Forest algorithm to learn taxonomical semantic relations and they have shown that the model performs well for co-hyponymy detection. In another attempt,~\newcite{JANA18.78} proposed various complex network measures which can be used as features to build a supervised classifier model for co-hyponymy detection, and showed improvements over other baseline approaches.  Recently, with the emergence of various network representation learning methods~\cite{perozzi2014deepwalk,tang2015line,grover2016node2vec,ribeiro2017struc2vec}, attempts have been made to convert distributional thesauri network into low dimensional vector space.~\cite{ferret2017turning}
apply distributional thesaurus embedding for synonym extraction and expansion tasks whereas ~\newcite{jana2018can} use it to improve the state-of-the-art performance of word similarity/relatedness tasks, word analogy task etc. 

Thus, a natural question arises as to whether network embeddings should be more effective than the handcrafted network features used by~\newcite{JANA18.78} for co-hyponymy detection. Being motivated by this connection, we investigate how the information captured by network representation learning methodologies on distributional thesaurus can be used in discriminating word pairs having co-hyponymy relation from the word pairs having hypernymy, meronymy relation or any random pair of words. We use the distributional thesaurus (DT) network~\cite{riedl2013scaling} built using Google books syntactic n-grams. As a network representation learning method, we apply node2vec~\cite{grover2016node2vec} which is an algorithmic framework for learning continuous feature representations for nodes in networks that maximizes the likelihood of preserving network
neighborhoods of nodes. Thus obtained vectors are then used as feature vectors and plugged into the classifiers according to the state-of-the-art experimental setup.  \\

\noindent {\bf Classification model:} To distinguish the word pairs having co-hyponymy relation from the word pairs having hypernymy or meronymy relation, or from any random pair of words, we combine the network embeddings of the two words by concatenation (CC) and addition (ADD) operations to provide as features to train classifiers like Support Vector Machine (SVM) and Random Forest (RF).\\

\noindent {\bf Evaluation results:} We evaluate the usefulness of DT embeddings against three benchmark datasets for co-hyponymy detection~\cite{weeds2014learning,SANTUS16.455,JANA18.78}, following their experimental setup. We show that the network embeddings outperform the baselines by a huge margin throughout all the experiments, except for co-hyponyms vs. random pairs, where the baselines already have very high accuracy and network embeddings are able to match the results. 

\section{Methodology}
\label{method}
We take the distributional thesaurus (DT)~\cite{riedl2013scaling} constructed from the Google books syntactic n-grams data~\cite{goldberg2013dataset} spanning from 1520 to 2008 as the underlying network where each word's neighborhood is represented by a list of top 200 words that are similar with respect to their bi-gram distribution~\cite{riedl2013scaling}. 
\begin{figure}[!ht]
\begin{center}
\includegraphics[scale=0.35]{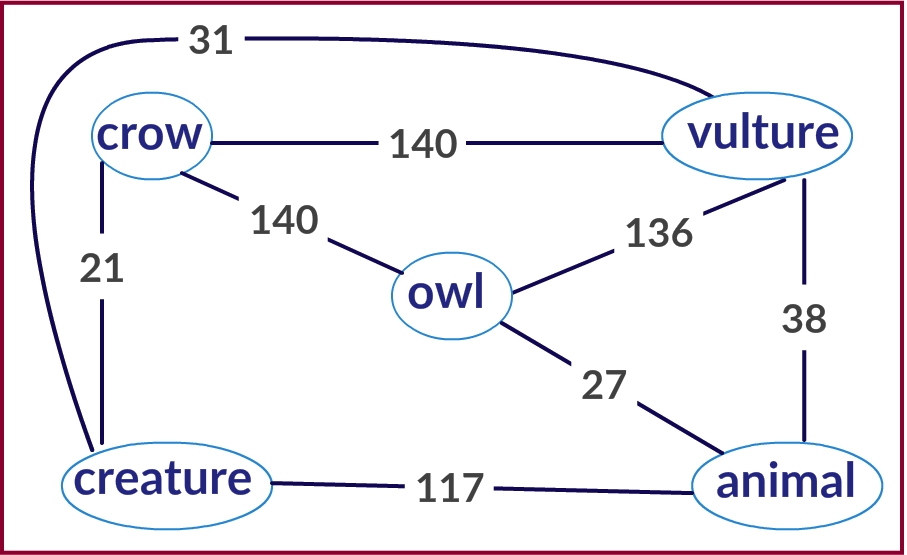} 
\caption{A sample snapshot of distributional thesaurus (DT) network, where each node represents a word and the weight of edge between two nodes is defined as the number of context features that these two words share in common. Here the word `owl' shares more context features with its co-hyponyms -- `crow', `vulture' compared to their hypernym `animal'.}
\label{DT}
\end{center}
\end{figure}

The nodes in the network represent words and edges are present between a node and its top 200 similar nodes; the number of features that two nodes share in common is assigned as the weight of the edge connecting them. A snapshot of the DT is shown in Figure~\ref{DT}. We see that a target word `owl' is connected with its co-hyponyms, `crow' and `vulture' via higher weighted edges, whereas the edge weights with its hypernyms like `animal' are less. It may also happen that hypernyms of a target word are not even present in its neighborhood. For example, `creature' is not present in the neighborhood of `owl' but it is connected with `crow' via less weighted edge. As per the DT network structure, distributionally similar words are present in a close proximity with similar neighborhood. \\

According to the literature dealing with lexical relation detection, words having co-hyponymy relation are distributionally more similar than the words having hypernymy or meronymy relation or any random pair of words. This is well captured by the DT. In a recent work, ~\newcite{JANA18.78} used network features extracted from the DT to detect co-hyponyms. In our approach, we attempt to use embeddings obtained through a network representation learning method such as node2vec~\cite{grover2016node2vec} when applied over the DT network. By choosing a flexible notion of a neighborhood and applying a biased random walk procedure, which efficiently explores diverse neighborhoods, node2vec learn representations for each node that organize nodes based on their network roles and/or communities. We use the default setup of node2vec; having walk-length 80, walks per node 10, window size 10 and dimension of vector 128. 

\begin{figure}[!tbh]
\centering
\includegraphics[width=0.5\textwidth]{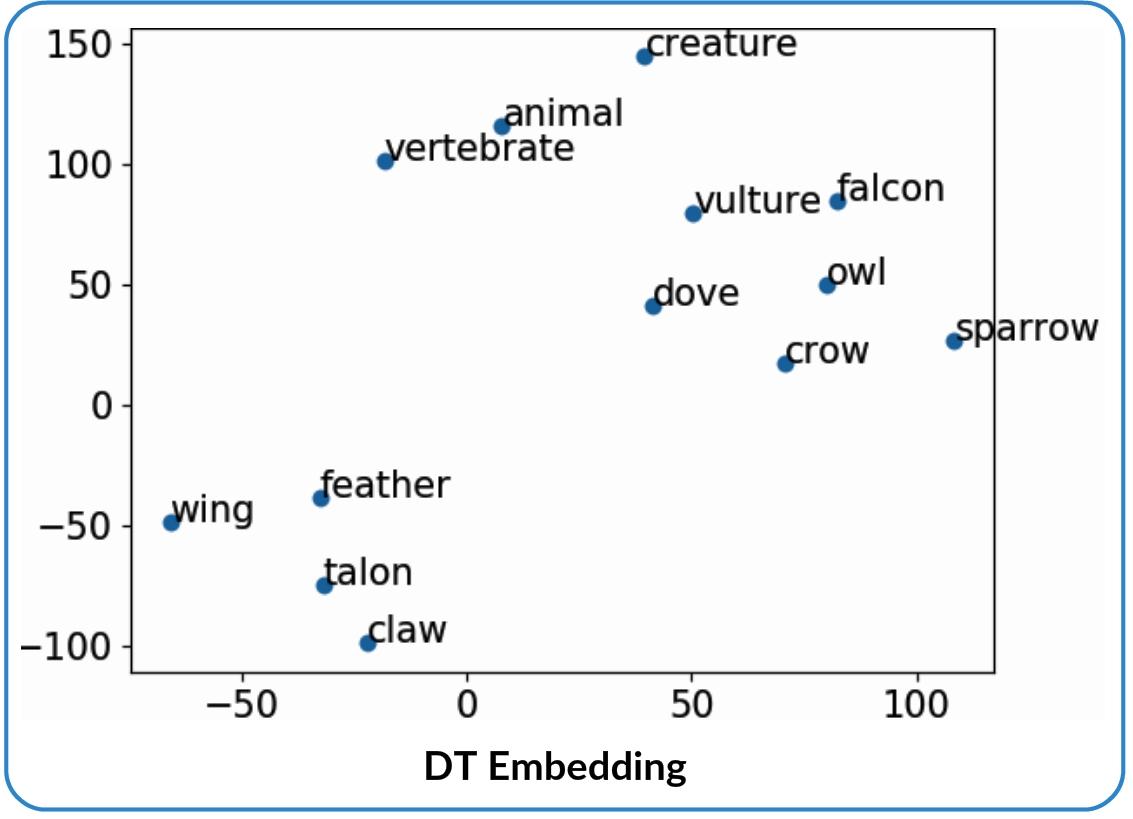}
\caption[t-Distributed Stochastic Neighbor (t-SNE)]{t-Distributed Stochastic Neighbor (t-SNE)~\cite{maaten2008visualizing} plot of DT embedding obtained using node2vec. 
}
\label{fig:induced}
\end{figure}

In order to do a qualitative analysis of the obtained vectors, we plot some sample words using t-SNE~\cite{maaten2008visualizing} 
in Figure~\ref{fig:induced}. We observe that the relative distance between the co-hyponymy pairs is much smaller than those having hypernymy relations or meronymy relations for the DT embeddings. For instance, the co-hyponyms of `owl' like `crow', `vulture', `sparrow' are close to each other whereas hypernyms of `owl' like `animal', `vertebrate', `creature', as well as meronyms of `owl' like `claw',`feather', are at distant positions. 

We aim to build a classifier that given a word pair, is able to detect whether or not they hold a co-hyponymy relation. Since we intend to explore the use of DT embeddings, 
we need to come up with specific ways to combine the embeddings of the word pair to be used as features for the classification. Following the literature~\cite{weeds2014learning}, we investigate four operations - vector difference (DIFF), vector concatenation (CC), vector pointwise addition (ADD) and vector pointwise multiplication (MUL). From our initial experiments, we find that CC and ADD prove to be the better combination methods overall. It is justified, as DIFF and MUL operations are somewhat intersective whereas both CC and ADD effectively come up with the union of the features in different ways and classifier fed with both shared and non-shared features has access to more information leading to better accuracy. 
We only report the performances for CC and ADD for Support Vector Machine (SVM) and Random Forest (RF) classifiers. 

\section{Experimental Results and Analysis}
\label{experiment}
We perform experiments using three benchmark datasets for co-hyponymy detection~\cite{weeds2014learning,SANTUS16.455,JANA18.78}. For each of these, we follow the same experimental setup as discussed by the authors and compare our method with the method proposed by the author as well as the state-of-the-art models by~\newcite{JANA18.78}. We perform the analysis of three datasets to investigate the extent of overlap present in these publicly available benchmark datasets and find out that 45.7\% word pairs of dataset prepared by~\newcite{weeds2014learning} are present in dataset ROOT9 prepared by~\newcite{SANTUS16.455}. This intersection set comprises 27.8\% of the ROOT9 dataset. Similarly  36.7\% word pairs of dataset prepared by~\newcite{weeds2014learning} are present in the whole dataset prepared by~\newcite{JANA18.78}. This intersection set comprises 44.9\% of the dataset prepared by~\newcite{JANA18.78}.
\begin{table}[!tbh] 
\centering
\begin{tabular}{|>{\centering}p{1.5cm}|>{\centering}p{5.6cm}|}
      \hline
       \textbf{Baseline Model}&\textbf{Description}\tabularnewline \hline
       svmDIFF& A linear SVM trained on the vector difference\tabularnewline\hline  
	   svmMULT& A linear SVM trained on the pointwise product vector\tabularnewline \hline
svmADD&A linear SVM trained on the vector sum\tabularnewline \hline 
svmCAT&A linear SVM trained on the vector concatenation\tabularnewline  \hline
svmSING&A linear SVM trained on the vector of the second word in the given word pair\tabularnewline \hline
 knnDIFF& $k$ nearest neighbours (knn) trained on the vector difference \tabularnewline\hline
       cosineP& The relation between word pair holds if the cosine similarity of the word vectors is greater than some threshold $p$\tabularnewline\hline  
       linP&The relation between word pair holds if the lin similarity~\cite{lin1998automatic} of the word vectors is greater than some threshold $p$\tabularnewline\hline
\end{tabular}
\caption{Descriptions of the baseline models as described in~\protect\cite{weeds2014learning}}
\label{Desc}
\end{table}

\begin{table}[!ht]
\centering
\begin{tabular}{|>{\centering}p{3.2cm}|>{\centering}p{2.4cm}|}
      \hline      \textbf{Model}&\textbf{Accuracy}\tabularnewline \hline
       \cellcolor{green}svmDIFF& \cellcolor{green}0.62\tabularnewline  

svmMULT&0.39\tabularnewline 
       svmADD&0.41\tabularnewline  
       svmCAT&0.40\tabularnewline  
       svmSING&0.40\tabularnewline \hline 
       knnDIFF&0.58\tabularnewline
       
       \cellcolor{green}cosineP&\cellcolor{green}0.79\tabularnewline  
       linP&0.78\tabularnewline
       \hline
	   
\end{tabular}
\caption{Accuracy scores on a ten-fold cross validation for $cohyponym_{BLESS}$ dataset of all the baseline models described in~\protect\cite{weeds2014learning}} 
\label{C_BASE}
\end{table}

\begin{table}[!ht]
\centering
\begin{tabular}{|>{\centering}p{3.5cm}|>{\centering}p{1.9cm}|>{\centering}p{1.4cm}|}
      \hline      &\textbf{Model}&\textbf{Accuracy}\tabularnewline \hline
       \multirow{2}{*}{~\cite{weeds2014learning}}&svmDIFF& 0.62\tabularnewline  
\if{0}
&svmMULT&0.39\tabularnewline 
       &svmADD&0.41\tabularnewline  
       &svmCAT&0.40\tabularnewline  
       &svmSING&0.40\tabularnewline 
       &knnDIFF&0.58\tabularnewline
       \fi
       &cosineP&0.79\tabularnewline  
       \hline
	    \multirow{1}{*}{~\cite{JANA18.78}}&svmSS&0.84\tabularnewline  
       \hline\hline
       	    \multirow{4}{*}{\textbf{Our models}}&SVM\_CC&0.84\tabularnewline  
       &SVM\_ADD&0.9 \tabularnewline
      &RF\_CC&\cellcolor{green}\textbf{0.97} \tabularnewline
       &RF\_ADD&0.95 \tabularnewline
\hline
\end{tabular}
\caption[Accuracy scores on a ten-fold cross validation for $cohyponym_{BLESS}$ dataset of our models along with the top two baseline models (one supervised, one semi-supervised) described  in]{Accuracy scores on a ten-fold cross validation for $cohyponym_{BLESS}$ dataset of our models along with the top two baseline models (one supervised, one semi-supervised) described in~\cite{weeds2014learning} and models described in~\cite{JANA18.78}} 
\label{C_BLESS}
\end{table}

\begin{table}
\centering
\begin{tabular}{|>{\centering}p{3.5cm}|>{\centering}p{1.6cm}|>{\centering}p{1.6cm}|} 
\hline
      \textbf{Method}& \textbf{Co-Hyp vs Random}& \textbf{Co-Hyp vs Hyper} \tabularnewline \hline
	   \cite{SANTUS16.455} & 97.8 & 95.7\tabularnewline \hline
       \cite{JANA18.78}&\cellcolor{green} \textbf{99.0}& 87.0\tabularnewline \hline \hline
SVM\_CC&96.5&91.4\tabularnewline \hline 
SVM\_ADD&93.5&97.6\tabularnewline \hline     
RF\_CC&\cellcolor{green}\textbf{99.0}&98.6\tabularnewline \hline 
RF\_ADD&97.03&\cellcolor{green}\textbf{99.0}\tabularnewline \hline 
\end{tabular}
\caption{Percentage F1 scores on a ten-fold cross validation of our models along with the best models described in~\protect\cite{SANTUS16.455} and~\protect\cite{JANA18.78} for ROOT9 dataset}
\label{ROOT9}

\end{table}

\subsection{Experiment-1~\protect\cite{weeds2014learning}} \newcite{weeds2014learning} prepared \boldmath$cohyponym_{BLESS}$ dataset from the BLESS dataset~\cite{baroni2011we}. \boldmath$cohyponym_{BLESS}$ contains 5,835 labeled pair of nouns; divided evenly into pairs having co-hyponymy relations and others (having hypernymy, meronymy relations along with random word pairs). In their work,~\newcite{weeds2014learning} represent each word as positive pointwise mutual information (PPMI) based feature vector and propose a set of baseline methodologies, the descriptions of which are presented in Table~\ref{Desc}.


Following the same experimental setup, we report the accuracy measure for ten-fold cross validation and compare our models with the baselines in proposed by~\newcite{weeds2014learning}. Table~\ref{C_BASE} represents the performance of all the baseline models proposed by~\newcite{weeds2014learning}. In Table~\ref{C_BLESS} we show the performance of the best supervised model (svmDIFF) and the best semi-supervised model (cosineP) proposed by~\newcite{weeds2014learning} along with our models.

Here, the best model proposed by ~\newcite{JANA18.78} uses SVM classifier which is fed with structural similarity of the words in the given word pair from the distributional thesaurus network. 
We see that all the 4 proposed methods perform at par or better than the baselines, and using RF\_CC gives a 15.4\% improvement over the best results reported. 

\subsection{Experiment-2~\protect\cite{SANTUS16.455}} In the second experiment, we use \textbf{ROOT9} dataset prepared by~\newcite{SANTUS16.455}, containing 9,600 labeled pairs extracted from three datasets: EVALution~\cite{santus2015evalution}, Lenci/Benotto~\cite{benotto2015distributional} and BLESS~\cite{baroni2011we}. There is an even distribution of the three classes (hypernyms, co-hyponyms and random) in the dataset. Following the same experimental setup as~\cite{SANTUS16.455}, we report percentage F1 scores on a ten-fold cross validation for binary classification of co-hyponyms vs random pairs, as well as co-hyponyms vs. hypernyms using both SVM and Random Forest classifiers. 
Table~\ref{ROOT9} represents the performance comparison of our models with the best state-of-the-art models reported in~\cite{SANTUS16.455} and ~\cite{JANA18.78}. Here, the best model proposed by~\newcite{SANTUS16.455} uses Random Forest classifier which is fed with nine corpus based features like frequency of words, co-occurrence frequency etc., and the best model proposed by~\newcite{JANA18.78} use Random Forest classifier which is fed with five complex network features like structural similarity, shortest path etc. computed from the distributional thesaurus network. The results in Table~\ref{ROOT9} shows that, for the binary classification task of co-hyponymy vs random pairs, we achieve percentage F1 score of 99.0 with RF\_CC which is at par with the state-of-the-art models. More importantly, both RF\_CC and RF\_ADD beat the baselines with significant margins for the classification task of co-hyponymy vs hypernymy pairs. 

\begin{table}[!ht]
\centering
\begin{tabular}{|>{\centering}p{1.7cm}|>{\centering}p{1.5cm}|>
{\centering}p{1.5cm}|>
{\centering}p{1.5cm}|}

      \hline
      \textbf{Model} & \textbf{Co-Hyp vs Random} & \textbf{Co-Hyp vs Mero}& \textbf{Co-Hyp vs Hyper}\tabularnewline
      \hline
       svmSS & 0.96 & 0.86&0.73\tabularnewline \hline
	   rfALL & 0.97 & 0.89&0.78\tabularnewline  \hline\hline
       SVM\_CC & 0.9 & 0.89 &0.854\tabularnewline  \hline
       SVM\_ADD & 0.943 & 0.89 &0.869\tabularnewline  \hline
       RF\_CC & 0.97 &\cellcolor{green}\textbf{0.978} &\cellcolor{green}\textbf{0.98}\tabularnewline  \hline
       RF\_ADD & \cellcolor{green}\textbf{0.971} & 0.956 &0.942\tabularnewline  \hline
\end{tabular}
\caption[Accuracy scores on a ten-fold cross validation of models (svmSS, rfALL) proposed by]{Accuracy scores on a ten-fold cross validation of models (svmSS, rfALL) proposed by~\newcite{JANA18.78} and our models for the dataset prepared by~\newcite{JANA18.78}.}
\label{PE}

\end{table}

\subsection{Experiment-3~\protect\cite{JANA18.78}} In the third experiment we use the dataset specifically build for co-hyponymy detection in one of the recent works by~\newcite{JANA18.78}. This dataset is extracted from BLESS~\cite{baroni2011we} and divided into three small datasets- \textbf{Co-Hypo vs Hyper, Co-Hypo vs Mero, Co-Hypo Vs Random}. Each of these datasets are balanced, containing 1,000 co-hyponymy pairs and 1,000 pairs for the other class. 
Following the same setup, we report accuracy scores for ten-fold cross validation for each of these three datasets of our models along with the best models (svmSS, rfALL) reported by~\newcite{JANA18.78} in Table~\ref{PE}.~\newcite{JANA18.78} use SVM classifier with structural similarity between words in a word pair as feature to obtain svmSS and use Random Forest classifier with five complex network measures computed from distributional thesaurus network as features to obtain rfALL. From the results presented in Table~\ref{PE}, RF\_CC proves to be the best among our proposed models which performs at par with the baselines for \textbf{Co-Hypo vs Random} dataset. Interestingly, it beats the baselines comprehensively for \textbf{Co-Hypo vs Mero} and \textbf{Co-Hypo vs Hyper} datasets, providing improvements of 9.88\% and 25.64\%, respectively.\\

\subsection{Error Analysis} We further analyze the cases for which our model produces wrong prediction. We point out some example word pairs such as `screw - screwdriver', `gorilla - orangutan' from \boldmath$cohyponym_{BLESS}$ dataset which our model wrongly flags as `false'. 
We observe a drastic difference in frequency between the words in these words pairs in the corpus from which the DT was constructed; for example `screw' appears 592,857 times whereas `screwdriver' has a frequency of 29,748; similarly `gorilla' has a frequency of 40,212 whereas `orangutan' has 3,567. In the DT network, edge weight depends on the overlap between top 1000 context features, and a drastic frequency difference might not capture this well. On the other hand, there are examples like `potato - peel', `jacket - zipper' which our model wrongly flags as `true' co-hyponyms. We observe that the corpus does not contain many co-hyponyms of `peel' or `zipper', and thus their neighborhood in the DT network contains words like `ginger, lemon, onion, garlic' and `pant, skirt, coat, jeans' which are co-hyponyms of `potato' and `jacket', respectively. This leads to the false signal by the approach. 

\if{0}
\noindent \textbf{Comparison with the state-of-the-art word representation: }
We conduct all the experiments discussed in Section~\ref{experiment} by plugging in GloVe vectors instead of DT embeddings and observe that for all the experiment settings, either DT embedding performs at par with GloVe or it outperforms GloVe. In addition to that, being motivated by~\newcite{jana2018can}, we also try the joint representation of DT embedding and GloVe. We find that the performance of the joint representation is very close to the performance of DT embeddings used alone. Please refer to the supplementary material for these results.
\fi
\section{Conclusion}
In this paper, we have investigated how the distributional thesaurus embeddings obtained using network representation learning can help improve the otherwise difficult task of discriminating co-hyponym pairs from hypernym, meronym and random pairs. By extensive experiments, we have shown that while the proposed models are at par with the baselines for detecting co-hyponyms vs. random pairs, they outperform the state-of-the-art models by a huge margin for the binary classification of co-hyponyms vs. hypernyms, as well as co-hyponyms vs. meronyms.  It clearly shows that network representations can be very effectively utilized for a focused task like relation extraction.  All the datasets, DT embeddings and codes (with instructions) used in our experiments are made publicly available\footnote{\url{https://tinyurl.com/u55np6o}}.\\

The next immediate step is to try out DT embedding to build unsupervised model for co-hyponymy detection. In future, we plan to investigate some more sophisticated network representation learning techniques like path embedding, community embedding techniques to embed the path joining the given pair of words or the subgraph induced by the given pair of words etc. and apply it on distributional thesaurus network for robust detection of lexical relations. In this study, our focus has been distinguishing a horizontal relation, co-hyponymy, from parent-child relations like hypernymy and meronymy. However, the investigation on discriminating two analogous sibling relations, co-hyponymy and co-meronymy using the proposed method would be one of the interesting future direction. Finally, our broad objective is to build a general supervised and unsupervised framework based on complex network theory to detect different lexical relations from a given a corpus with high accuracy.

\section{Acknowledgements}

This research was supported by the Deutsche Forschungsgemeinschaft (DFG) under the project ``Joining Ontologies and Semantics Induced from Text" (JOIN-T 2, BI 1544/4-2, PO 1900/1-2).

\if{0}
\subsection{General Instructions for the Submitted Full Paper}

Each submitted paper  should be between \ul{a minimum of four  and
a maximum of eight  pages including figures}.

\section{Full Paper}

Each manuscript should be submitted on white A4 paper. The fully
justified text should be formatted in two parallel columns, each 8.25 cm wide,
and separated by a space of 0.63 cm. Left, right, and bottom margins should be
1.9 cm. and the top margin 2.5 cm. The font for the main body of the text should
be Times New Roman 10 with interlinear spacing of 12 pt.  Articles must be
between 4 and 8 pages in length, regardless of the mode of presentation (oral
or poster).

\subsection{General Instructions for the Final Paper}

Each paper is allocated between \ul{a minimum of four and a maximum of
eight pages including figures}. The unprotected PDF files will appear in the
on-line proceedings directly as received. Do not print the page number.

\section{Page Numbering}

\textbf{Please do not include page numbers in your article.} The definitive page
numbering of articles published in the proceedings will be decided by the
organising committee.

\section{Headings / Level 1 Headings}

Headings should be capitalised in the same way as the main title, and centred
within the column. The font used is Times New Roman 12 bold. There should
also be a space of 12 pt between the title and the preceding section, and
a space of 3 pt between the title and the text following it.

\subsection{Level 2 Headings}

The format for level 2 headings is the same as for level 1 Headings, with the
font Times New Roman 11, and the heading is justified to the left of the column.
There should also be a space of 6 pt between the title and the preceding
section, and a space of 3 pt between the title and the text following it.

\subsubsection{Level 3 Headings}

The format for level 3 headings is the same as for level 2 headings, except that
the font is Times New Roman 10, and there should be no space left between the
heading and the text. There should also be a space of 6 pt between the title and
the preceding section, and a space of 3 pt between the title and the text
following it.

%

\section{Citing References in the Text}

\subsection{Bibliographical References}

All bibliographical references within the text should be put in between
parentheses with the author's surname followed by a comma before the date
of publication,\cite{Martin-90}. If the sentence already includes the author's
name, then it is only necessary to put the date in parentheses:
\newcite{Martin-90}. When several authors are cited, those references should be
separated with a semicolon: \cite{Martin-90,CastorPollux-92}. When the reference
has more than three authors, only cite the name of the first author followed by
``et al.'' (e.g. \cite{Superman-Batman-Catwoman-Spiderman-00}).

\subsection{Language Resource References}

\subsubsection{When Citing Language Resources}

When citing language resources, we recommend to proceed in the same way to
bibliographical references.
Thus, a language resource should be cited as \cite{speecon}.

\section{Figures \& Tables}
\subsection{Figures}

All figures should be centred and clearly distinguishable. They should never be
drawn by hand, and the lines must be very dark in order to ensure a high-quality
printed version. Figures should be numbered in the text, and have a caption in
Times New Roman 10 pt underneath. A space must be left between each figure and
its respective caption. 

Example of a figure enclosed in a box:

\begin{figure}[!h]
\begin{center}
\includegraphics[scale=0.5]{lrec2020W-image1.eps} 
\caption{The caption of the figure.}
\label{fig.1}
\end{center}
\end{figure}

Figure and caption should always appear together on the same page. Large figures
can be centred, using a full page.

\subsection{Tables}

The instructions for tables are the same as for figures.
%
\begin{table}[!h]
\begin{center}
\begin{tabularx}{\columnwidth}{|l|X|}

      \hline
      Level&Tools\\
      \hline
      Morphology & Pitrat Analyser\\
      \hline
      Syntax & LFG Analyser (C-Structure)\\
      \hline
     Semantics & LFG F-Structures + Sowa's\\
     & Conceptual Graphs\\
      \hline

\end{tabularx}
\caption{The caption of the table}
 \end{center}
\end{table}

%
%
%
%
%

\section{Footnotes}

Footnotes are indicated within the text by a number in
superscript\footnote{Footnotes should be in Times New Roman 9 pt, and appear at
the bottom of the same page as their corresponding number. Footnotes should also
be separated from the rest of the text by a horizontal line 5 cm long.}.

\section{Copyrights}

The Language Resouces and Evaluation Conference (LREC)
proceedings are published by the European Language Resources Association (ELRA).
They are available online from the conference website.

ELRA's policy is to acquire copyright for all LREC contributions. In assigning
your copyright, you are not forfeiting your right to use your contribution
elsewhere. This you may do without seeking permission and is subject only to
normal acknowledgement to the LREC proceedings. The LREC 2020 Proceedings are
licensed under CC-BY-NC, the Creative Commons Attribution-Non-Commercial 4.0
International License.

\section{Conclusion}

Your submission of a finalised contribution for inclusion in the LREC
proceedings automatically assigns the above-mentioned copyright to ELRA.

\section{Acknowledgements}

Place all acknowledgements (including those concerning research grants and
funding) in a separate section at the end of the article.

\section{Providing References}

\subsection{Bibliographical References}
Bibliographical references should be listed in alphabetical order at the
end of the article. The title of the section, ``Bibliographical References'',
should be a level 1 heading. The first line of each bibliographical reference
should be justified to the left of the column, and the rest of the entry should
be indented by 0.35 cm.

The examples provided in Section \secref{main:ref} (some of which are fictitious
references) illustrate the basic format required for articles in conference
proceedings, books, journal articles, PhD theses, and chapters of books.

\subsection{Language Resource References}

Language resource references should be listed in alphabetical order at the end
of the article.

\section*{Appendix: How to Produce the \texttt{.pdf} Version}

In order to generate a PDF file out of the LaTeX file herein, when citing
language resources, the following steps need to be performed:

\begin{itemize}
    \item{Compile the \texttt{.tex} file once}
    \item{Invoke \texttt{bibtex} on the eponymous \texttt{.aux} file}
    \item{Compile the \texttt{.tex} file twice}
\end{itemize}
\fi
\section{Bibliographical References}
\label{main:ref}

\bibliographystyle{lrec}

\begin{thebibliography}{}

\bibitem[\protect\citename{Baroni and Lenci}2011]{baroni2011we}
Baroni, M. and Lenci, A.
\newblock (2011).
\newblock How we blessed distributional semantic evaluation.
\newblock In {\em Proceedings of the GEMS 2011 Workshop on GEometrical Models
  of Natural Language Semantics}, pages 1--10. Association for Computational
  Linguistics.

\bibitem[\protect\citename{Berland and Charniak}1999]{berland1999finding}
Berland, M. and Charniak, E.
\newblock (1999).
\newblock Finding parts in very large corpora.
\newblock In {\em Proceedings of the 37th annual meeting of the Association for
  Computational Linguistics on Computational Linguistics}, pages 57--64.
  Association for Computational Linguistics.

\bibitem[\protect\citename{Ferret}2017]{ferret2017turning}
Ferret, O.
\newblock (2017).
\newblock Turning distributional thesauri into word vectors for synonym
  extraction and expansion.
\newblock In {\em Proceedings of the Eighth International Joint Conference on
  Natural Language Processing (Volume 1: Long Papers)}, volume~1, pages
  273--283.

\bibitem[\protect\citename{Fu \bgroup et al.\egroup }2014]{fu2014learning}
Fu, R., Guo, J., Qin, B., Che, W., Wang, H., and Liu, T.
\newblock (2014).
\newblock Learning semantic hierarchies via word embeddings.
\newblock In {\em Proceedings of the 52nd Annual Meeting of the Association for
  Computational Linguistics (Volume 1: Long Papers)}, pages 1199--1209.

\bibitem[\protect\citename{Geffet and Dagan}2005]{geffet2005distributional}
Geffet, M. and Dagan, I.
\newblock (2005).
\newblock The distributional inclusion hypotheses and lexical entailment.
\newblock In {\em Proceedings of the 43rd Annual Meeting on Association for
  Computational Linguistics}, pages 107--114. Association for Computational
  Linguistics.

\bibitem[\protect\citename{Girju \bgroup et al.\egroup
  }2006]{girju2006automatic}
Girju, R., Badulescu, A., and Moldovan, D.
\newblock (2006).
\newblock Automatic discovery of part-whole relations.
\newblock {\em Computational Linguistics}, 32(1):83--135.

\bibitem[\protect\citename{Goldberg and Orwant}2013]{goldberg2013dataset}
Goldberg, Y. and Orwant, J.
\newblock (2013).
\newblock A dataset of syntactic-ngrams over time from a very large corpus of
  english books.
\newblock In {\em Second Joint Conference on Lexical and Computational
  Semantics (* SEM)}, volume~1, pages 241--247.

\bibitem[\protect\citename{Grover and Leskovec}2016]{grover2016node2vec}
Grover, A. and Leskovec, J.
\newblock (2016).
\newblock node2vec: Scalable feature learning for networks.
\newblock In {\em Proceedings of the 22nd ACM SIGKDD international conference
  on Knowledge discovery and data mining}, pages 855--864. ACM.

\bibitem[\protect\citename{Jana and Goyal}2018a]{jana2018can}
Jana, A. and Goyal, P.
\newblock (2018a).
\newblock Can network embedding of distributional thesaurus be combined with
  word vectors for better representation?
\newblock In {\em Proceedings of the 2018 Conference of the North American
  Chapter of the Association for Computational Linguistics: Human Language
  Technologies, Volume 1 (Long Papers)}, pages 463--473.

\bibitem[\protect\citename{Jana and Goyal}2018b]{JANA18.78}
Jana, A. and Goyal, P.
\newblock (2018b).
\newblock Network features based co-hyponymy detection.
\newblock In {\em Proceedings of the Eleventh International Conference on
  Language Resources and Evaluation (LREC 2018)}. European Language Resources
  Association (ELRA).

\bibitem[\protect\citename{Kiela \bgroup et al.\egroup
  }2015]{kiela2015exploiting}
Kiela, D., Rimell, L., Vulic, I., and Clark, S.
\newblock (2015).
\newblock Exploiting image generality for lexical entailment detection.
\newblock In {\em Proceedings of the 53rd Annual Meeting of the Association for
  Computational Linguistics (ACL 2015)}, pages 119--124. ACL.

\bibitem[\protect\citename{Lin}1998]{lin1998automatic}
Lin, D.
\newblock (1998).
\newblock Automatic retrieval and clustering of similar words.
\newblock In {\em Proceedings of the 17th international conference on
  Computational linguistics-Volume 2}, pages 768--774. Association for
  Computational Linguistics.

\bibitem[\protect\citename{Maaten and Hinton}2008]{maaten2008visualizing}
Maaten, L. v.~d. and Hinton, G.
\newblock (2008).
\newblock Visualizing data using t-sne.
\newblock {\em Journal of machine learning research}, 9(Nov):2579--2605.

\bibitem[\protect\citename{Morlane-Hond{\`e}re}2015]{morlane2015can}
Morlane-Hond{\`e}re, F.
\newblock (2015).
\newblock What can distributional semantic models tell us about part-of
  relations?
\newblock In {\em NetWordS}, pages 46--50.

\bibitem[\protect\citename{Nguyen \bgroup et al.\egroup
  }2017]{nguyen-EtAl:2017:EMNLP2017}
Nguyen, K.~A., K\"{o}per, M., Schulte~im Walde, S., and Vu, N.~T.
\newblock (2017).
\newblock Hierarchical embeddings for hypernymy detection and directionality.
\newblock In {\em Proceedings of the 2017 Conference on Empirical Methods in
  Natural Language Processing}, pages 233--243, Copenhagen, Denmark, September.
  Association for Computational Linguistics.

\bibitem[\protect\citename{Pantel and Pennacchiotti}2006]{pantel2006espresso}
Pantel, P. and Pennacchiotti, M.
\newblock (2006).
\newblock Espresso: Leveraging generic patterns for automatically harvesting
  semantic relations.
\newblock In {\em Proceedings of the 21st International Conference on
  Computational Linguistics and the 44th annual meeting of the Association for
  Computational Linguistics}, pages 113--120. Association for Computational
  Linguistics.

\bibitem[\protect\citename{Perozzi \bgroup et al.\egroup
  }2014]{perozzi2014deepwalk}
Perozzi, B., Al-Rfou, R., and Skiena, S.
\newblock (2014).
\newblock Deepwalk: Online learning of social representations.
\newblock In {\em Proceedings of the 20th ACM SIGKDD international conference
  on Knowledge discovery and data mining}, pages 701--710. ACM.

\bibitem[\protect\citename{Ribeiro \bgroup et al.\egroup
  }2017]{ribeiro2017struc2vec}
Ribeiro, L.~F., Saverese, P.~H., and Figueiredo, D.~R.
\newblock (2017).
\newblock struc2vec: Learning node representations from structural identity.
\newblock In {\em Proceedings of the 23rd ACM SIGKDD International Conference
  on Knowledge Discovery and Data Mining}, pages 385--394. ACM.

\bibitem[\protect\citename{Riedl and Biemann}2013]{riedl2013scaling}
Riedl, M. and Biemann, C.
\newblock (2013).
\newblock Scaling to large$^3$ data: An efficient and effective method to
  compute distributional thesauri.
\newblock In {\em Proceedings of the 2013 Conference on Empirical Methods in
  Natural Language Processing}, pages 884--890.

\bibitem[\protect\citename{Roller and Erk}2016]{roller-erk:2016:EMNLP2016}
Roller, S. and Erk, K.
\newblock (2016).
\newblock Relations such as hypernymy: Identifying and exploiting hearst
  patterns in distributional vectors for lexical entailment.
\newblock In {\em Proceedings of the 2016 Conference on Empirical Methods in
  Natural Language Processing}, pages 2163--2172, Austin, Texas, November.
  Association for Computational Linguistics.

\bibitem[\protect\citename{Roller \bgroup et al.\egroup
  }2014]{roller2014inclusive}
Roller, S., Erk, K., and Boleda, G.
\newblock (2014).
\newblock Inclusive yet selective: Supervised distributional hypernymy
  detection.
\newblock In {\em Proceedings of COLING 2014, the 25th International Conference
  on Computational Linguistics: Technical Papers}, pages 1025--1036.

\bibitem[\protect\citename{Santus \bgroup et al.\egroup
  }2014]{santus2014chasing}
Santus, E., Lenci, A., Lu, Q., and Im~Walde, S.~S.
\newblock (2014).
\newblock Chasing hypernyms in vector spaces with entropy.
\newblock In {\em Proceedings of the 14th Conference of the European Chapter of
  the Association for Computational Linguistics, volume 2: Short Papers}, pages
  38--42.

\bibitem[\protect\citename{Santus \bgroup et al.\egroup
  }2015]{santus2015evalution}
Santus, E., Yung, F., Lenci, A., and Huang, C.-R.
\newblock (2015).
\newblock Evalution 1.0: an evolving semantic dataset for training and
  evaluation of distributional semantic models.
\newblock In {\em Proceedings of the 4th Workshop on Linked Data in Linguistics
  (LDL-2015)}, pages 64--69.

\bibitem[\protect\citename{Santus \bgroup et al.\egroup }2016]{SANTUS16.455}
Santus, E., Lenci, A., Chiu, T.-S., Lu, Q., and Huang, C.-R.
\newblock (2016).
\newblock Nine features in a random forest to learn taxonomical semantic
  relations.
\newblock In {\em Proceedings of the Tenth International Conference on Language
  Resources and Evaluation (LREC 2016)}, Paris, France, may. European Language
  Resources Association (ELRA).

\bibitem[\protect\citename{Shwartz \bgroup et al.\egroup
  }2017]{shwartz-santus-schlechtweg:2017:EACLlong}
Shwartz, V., Santus, E., and Schlechtweg, D.
\newblock (2017).
\newblock Hypernyms under siege: Linguistically-motivated artillery for
  hypernymy detection.
\newblock In {\em Proceedings of the 15th Conference of the European Chapter of
  the Association for Computational Linguistics: Volume 1, Long Papers}, pages
  65--75, Valencia, Spain, April. Association for Computational Linguistics.

\bibitem[\protect\citename{Tang \bgroup et al.\egroup }2015]{tang2015line}
Tang, J., Qu, M., Wang, M., Zhang, M., Yan, J., and Mei, Q.
\newblock (2015).
\newblock Line: Large-scale information network embedding.
\newblock In {\em Proceedings of the 24th International Conference on World
  Wide Web}, pages 1067--1077. International World Wide Web Conferences
  Steering Committee.

\bibitem[\protect\citename{Weeds \bgroup et al.\egroup
  }2014]{weeds2014learning}
Weeds, J., Clarke, D., Reffin, J., Weir, D., and Keller, B.
\newblock (2014).
\newblock Learning to distinguish hypernyms and co-hyponyms.
\newblock In {\em Proceedings of COLING 2014, the 25th International Conference
  on Computational Linguistics: Technical Papers}, pages 2249--2259. Dublin
  City University and Association for Computational Linguistics.

\bibitem[\protect\citename{Yu \bgroup et al.\egroup }2015]{yu2015learning}
Yu, Z., Wang, H., Lin, X., and Wang, M.
\newblock (2015).
\newblock Learning term embeddings for hypernymy identification.
\newblock In {\em Twenty-Fourth International Joint Conference on Artificial
  Intelligence}.

\end{thebibliography}


\end{document}